\documentclass[10pt,twocolumn,letterpaper]{article}
\usepackage{graphicx}                       
\usepackage{graphics}                       
\usepackage{epsfig}                         
\usepackage[tight,footnotesize]{subfigure}  
\usepackage{amssymb,amsmath,amsfonts,amsthm,bm}
\usepackage{mdwmath}
\usepackage{commath}   
\usepackage{eqparbox}
\usepackage{mathtools}
\usepackage[utf8]{inputenc} 
\usepackage[english]{babel}
\usepackage{amsmath,amsfonts,amssymb}

\newcommand{\ie}{{\em i.e.\ }}

\newcommand{\eg}{{\em e.g.\ }}

\newcommand{\et}{{\em et al.\ }}

\newcommand{\LC}{\mbox{$\mathcal L$}}

\newcommand{\RC}{\mbox{$\mathcal R$}}

\newcommand{\beq}{\begin{equation}}
\newcommand{\eeq}{\end{equation}}
\newcommand{\bear}{\begin{eqnarray}}
\newcommand{\bears}{\begin{eqnarray*}}
\newcommand{\eear}{\end{eqnarray}}
\newcommand{\eears}{\end{eqnarray*}}
\newcommand{\bdm}{\begin{displaymath}}
\newcommand{\edm}{\end{displaymath}}
\newcommand{\lba}{\left[\begin{array}}
\newcommand{\ear}{\end{array}\right]}

\usepackage{stfloats}                       
\usepackage{url}
\usepackage{cite}                           
\usepackage[T1]{fontenc} 
\usepackage{tabularx}
\usepackage{multirow}
\usepackage[table]{xcolor}
\usepackage{longtable}
\usepackage{booktabs} 			
\usepackage{xcolor,colortbl}    

\graphicspath{ {images/} }
\usepackage{cvpr}
\usepackage{times}
\usepackage[pagebackref=true,breaklinks=true,letterpaper=true,colorlinks,bookmarks=false]{hyperref} 
\usepackage{authblk}
\cvprfinalcopy 
\def\cvprPaperID{2453} 

\ifcvprfinal\fi

\begin{document}
\title{Pose-based Modular Network for
Human-Object Interaction Detection}
\author{Zhijun Liang\textsuperscript{1}}
\author{Junfa Liu\textsuperscript{1}}
\author{Yisheng Guan\textsuperscript{1}}
\author{Juan Rojas\textsuperscript{2}}
\affil{\textsuperscript{1} Guangdong University of Technology \\
       \textsuperscript{2} Chinese University of Hong Kong}
\maketitle
\begin{abstract}
Human-object interaction(HOI) detection is a critical tasks in scene understanding. The goal is to infer the triplet $<$subject, predicate, object$>$ in a scene.   
In this work we note that the human pose itself as well as the relative spatial information of the human pose with respect to the target object can provide informative cues for HOI detection.
We contribute a Pose-based Modular Network (PMN) which explores the absolute pose features and relative spatial pose features to improve HOI detection and is fully compatible with existing networks. Our module consists of a branch that first processes the relative spatial pose features of each joint independently. Another branch updates the absolute pose features via fully-connected graph structures. The processed pose features are then fed into an action classifier. To evaluate our proposed method, we combine the module with the state-of-the-art model named VS-GATs and obtain significant improvement on two public benchmarks: V-COCO and HICO-DET, which shows its efficacy and flexibility.
Code is available at \url{https://github.com/birlrobotics/PMN}. 
\end{abstract}

\begin{figure}[t]
  \begin{center}
      \includegraphics[width=1\linewidth]{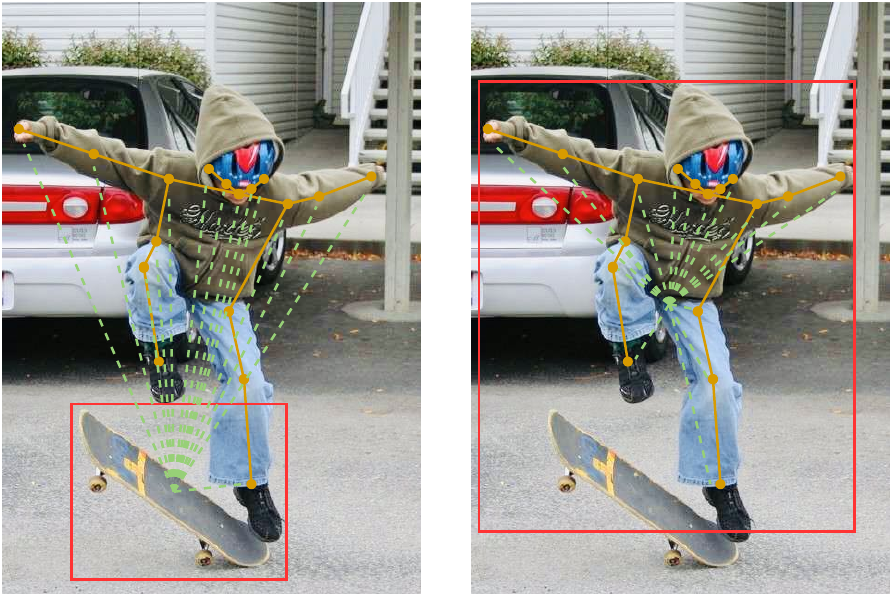}
  \end{center}
      \caption{
        \textbf{Two constructed pose features we use in our method}. The \textit{relative spatial pose features} (left) are the offset between each joint of human pose and the target object, which provides more detailed spatial information. The \textit{absolute pose features} (right) are the normalized keypoint features with respect to the human bounding box, which offers the pose intrinsic properties cues to the model.
      }
  \label{fig:insight_picture}
\end{figure}

\section{Introduction}\label{sec:intro}
Recently, great progress has been made in computer vision, including object detection \cite{renNIPS15fasterrcnn,liu2016ssd,dai2016r,girshick2015fast}, human pose estimation \cite{Dabral2018LearningMotion,pavllo20193d,liu2020gastnet,Zhao_2019}, action recognition \cite{Yan2018SpatialRecognition,Li2019Actional-StructuralRecognition} and scene segmentation \cite{he2017mask}. However, to better understand the visual world, a robot should not only detect the individual instances in a scene but also further comprehend how a person interact with the world. A subclass of that interaction is with objects. As such, human-object interaction (HOI) detection
has recently attracted increasing attention in the field of computer vision.

Human-object interaction detection infers the triplet \textit{$<$subject, predicate, object$>$} in a scene. For example, in Fig. \ref{fig:insight_picture}, we first detect the \textit{human} and object (\textit{skateboard}) instances. We finally infer the interaction \textit{ride} between them, yielding the triplet \textit{$<$human, ride, skateboard$>$}. Note that some images may contain multiple humans simultaneously interacting with various objects. One person may also have different interactions with a single object. For instance, Fig. \ref{fig:insight_picture} contains the set of ground-truth triplets: \textit{$<$human, ride, skateboard$>$}, \textit{$<$human, jump, skateboard$>$} and \textit{$<$human, stand\_on, skateboard$>$}.

Recently, researchers have proposed a variety of networks for HOI detection \cite{chao2018learning, gkioxari2018detecting, gao2018ican,gupta2018nofrills,Li_2019_CVPR,liang2020visualsemantic}. The first works were multi-stream neural networks that leveraged visual and spatial cues for HOI detection \cite{chao2018learning, gkioxari2018detecting, gao2018ican}.  Others have considered human pose or human part features and have outperformed previous works by a great margin showing that HOI detection system benefit from relevant context \cite{gupta2018nofrills, Li_2019_CVPR, wan2019pose}. More recently, Liang \et \cite{liang2020visualsemantic} propose a dual-graph attention network which enables the model to leverage the rich information by integrating and broadcasting information through the graph structure. However, they don't consider the useful human pose cues. 

In this paper, we study fine-grained human poses via relative and absolute pose features (Fig. \ref{fig:insight_picture}) to aid HOI detection. The models receives detailed spatial information in the form of relative spatial pose features between each human's keypoint coordinates (\ie the joint) and the center of the target object bounding box. Moreover, the human pose intrinsic properties can also provide useful cues. For example, \textit{$<$human, eat, apple$>$} and \textit{$<$human, drink\_with, bottle$>$} may have the similar posture. So we also use the absolute pose features, which consist of the keypoint coordinates normalized to the center of the human bounding box. 

Furthermore, we propose a Pose-based Modular Network (PMN) which explores the constructed pose features (Fig. \ref{fig:insight_picture}) and is fully compatible with existing networks for HOI detection. The module consists of one branch that processes the relative spatial pose features of each joint independently and another branch which uses graph convolutions to update the absolute pose features of each joint. We then fuse the processed features followed by an action classifier as depicted in Fig. \ref{fig:overview}.

We evaluate our proposed module on two public benchmarks V-COCO \cite{DBLP:GuptaM15} and HICO-DET \cite{chao2018learning}. Our method consistently improves the state-of-the-art method \cite{liang2020visualsemantic}. On V-COCO, our method improves SOTA by \textbf{2 mAP} ($\sim$4.0\%). On the more challenging HICO-DET, our method improves SOTA by \textbf{0.98 mAP} ($\sim$4.6\%), \textbf{1.57 mAP} ($\sim$9.8\%), \textbf{0.75 mAP} ($\sim$3.5\%) for the Full, Rare and Non-Rare categories respectively. The addition of human pose cues to visual, spatial, and semantic cues; whilst being attended with attention, aided to further reduce false positives in the crowded scenes in general (Fig. \ref{fig:qualitative}). The improvements indicate our method is efficient and flexible.

\section{Related work}\label{sec:relatedWorks}

\begin{figure*}
    \centering
    \includegraphics[width=1\linewidth]{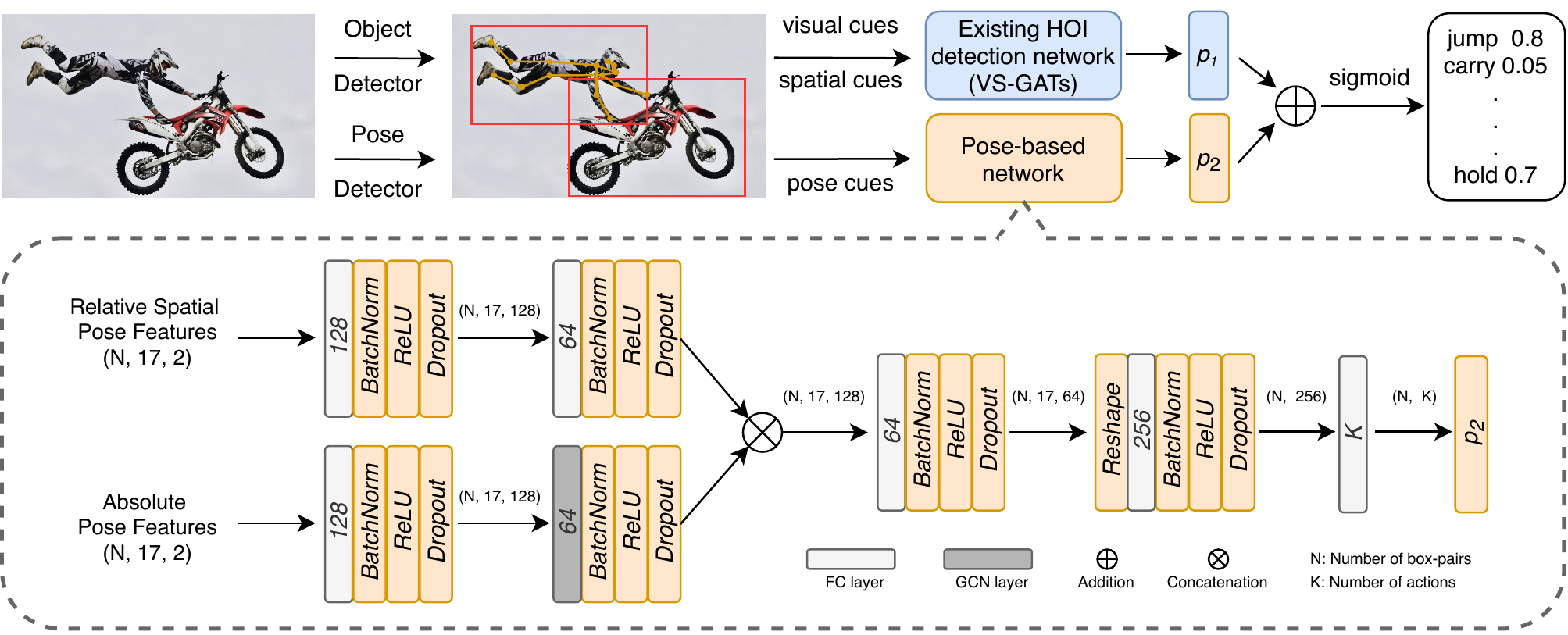}
    \caption{\textbf{Framework Overview}. Our system consists of two streams: a) an existing HOI detection network for inference based on supplied cues \textit{(e.g.} visual, spatial, semantic.); b) our proposed pose-based modular network that extends the top branch for better contextualization with absolute and relative pose cues. 
    The bottom half of the diagram depicts the pose-based network design in detail.}
    \label{fig:overview}
\end{figure*}

\paragraph{\textbf{Object Detection and Pose Estimation.}}
In scene understanding, object detection \cite{renNIPS15fasterrcnn,liu2016ssd,dai2016r,girshick2015fast} identifies, localizes, and classifies object instances in a scene.
Pose estimation \cite{Dabral2018LearningMotion,pavllo20193d,liu2020gastnet} 
computes 2D or 3D coordinates of human skeleton keypoints (body joints like shoulders, eyes, and knees often 17 in total). For HOI detection, researchers have used off-the-shelf object detector to localize people and objects. They have also adopted pose estimators to obtain fine-grained human poses. Then, instances and human pose features are leveraged by neural architectures for HOI inference.
\paragraph{\textbf{Graph Neural Network.}}
Graph neural networks (GNN) \cite{Wu2019a,niepert2016learning,velivckovic2017graph,kipf2016semisupervised,xu2018how,hamilton2017inductive} have recently grained increasing attention. Kipf \et \cite{kipf2016semisupervised} proposed a variant of graph convolutional networks (GCN) by introducing a first-order approximation to spectral graph convolutions. Velivckovic \et \cite{velivckovic2017graph} introduced graph attention networks (GATs) which leveraged masked self-attentional layers to enable its nodes to attend their neighborhood features with varying dynamic weights. Lately, GNNs have been successfully applied to pose estimation \cite{liu2020gastnet,Zhao_2019}. Inspired by \cite{liu2020gastnet,Zhao_2019}, we also use GCNs to encode absolute pose features.
\paragraph{\textbf{Human-Object Interaction Detection.}}
Improving HOI detection requires the model to better leverage contextual information in complex scenes. Chao \et \cite{chao2018learning} contributed the HICO-DET dataset and proposed a novel DNN input named Interaction Pattern to represent the spatial relations. Gkioxari \et \cite{gkioxari2018detecting} designed InteractNet that predicts a density over target object locations based on the appearance of a detected person. Gao \et \cite{gao2018ican} extended the methods in \cite{chao2018learning,gkioxari2018detecting} by introducing an instance-centric attention module to dynamically highlight the region of interest in an image. Different from these multi-stream neural networks, Qi \et \cite{qi2018learning} introduced the Graph Parsing Neural Network (GPNN) which iteratively updates features over a graph structure. Recently, Xu \et \cite{xu2019learning} considered the intrinsic semantic regularities across the scene to facilitate HOI detection. Liang \et \cite{liang2020visualsemantic} contributed a dual-graph attention network which takes visual, spatial and semantic cues to learn rich relations across scene instances over the novel graph network. Li \et \cite{Li_2019_CVPR} and Wan \et \cite{wan2019pose} further combine the fine-grained human pose and interaction pattern \cite{chao2018learning} as the spatial configuration map followed by a MLP and later concatenate all the processed features from multiple branches. However, this design can not be fully transferred to existing networks. Lately, Gupta \et \cite{gupta2018nofrills} designed a factored model which considered human appearance features, boxed-pair configurations, and fine-grained human poses as isolated factors for HOI detection. However, in their fine-grained layout factor network, they simply flatten and concatenate the pose information (including relative and absolute pose features) and feed them into an MLP. In our method, we first process the relative and absolute pose features separately with different networks and later fuse and flatten them before feeding into the classifier (Fig. \ref{fig:overview}). Our proposed pose-based module is fully compatible with existing networks and yields a significant gain in performance.

\section{Method}\label{sec:method}  
In this section,  we start with an overview of the entire system (Sec. \ref{subsec:overview}) followed by an introduction to the various pose features considered in our work (Sec. \ref{subsec:relatedFeatures}). Then, we outline our pose-based modular neural network structure (Sec. \ref{subsec:PMN}). Finally, we describe inference and learning for our model (Sec. \ref{subsec:inference}).
\subsection{Overview}\label{subsec:overview} 
As illustrated in Fig. \ref{fig:overview}, our system consists of two branches. Given an input image: (i) an off-the-shelf object detector \cite{renNIPS15fasterrcnn} extracts instance bounding boxes along with their embedded features and (ii) an off-the-shelf pose detector \cite{he2017mask} extracts the human pose keypoints. Suitable features are constructed and fed both into the existing visual-semantic graph attention HOI detection network and the proposed pose-based network. Score factors (the output of last layer of each stream) $p_1$ and $p_2$ are generated, summed and fed into a sigmoid function to predict the score for each action/predicate.

Specifically, for each human-object pair, we denote $s_h$ and $s_o$ as the confidence scores of the detected human and object instances respectively. We denote $s^a$ as the score of action $a \in \{1,...,K\}$, where \textit{K} is the total number of possible actions. The final score of the HOI triplet \textit{$<$subject, predicate, object$>$} is the product of the scores:
\begin{equation}
    \label{Eqtn:triple_score}
    S = s_h * s_o * s^a .
\end{equation}
We choose VS-GATs \cite{liang2020visualsemantic} as the existing HOI detection network in our framework as it has shown that by capturing visual-spatial and semantic cues via independent attention mechanisms that are later combined, the network is able better disambiguate hard detection cases and beating the state-of-the-art on the challenging HICO-DET dataset and ranking second place on the small-scale V-COCO dataset. Additionally, their code and preprocessed features are publicly available which enables testing and benchmarking for the community.  
\subsection{Pose Features} \label{subsec:relatedFeatures}
\subsubsection{Relative Spatial Pose Features}\label{subsubsec:rel_pose_feat} 
Spatial features are able to provide informative cues to infer the predicate. For example, the \textit{human} box above the \textit{skateboard} box strongly indicates the \textit{ride} interaction. 
In past works, there have been two main approaches to encode spatial relationship between person and object \cite{chao2018learning,gao2018ican,Li_2019_CVPR,wan2019pose}. 
Works like those of Chao \et \cite{chao2018learning}, adopt an interaction representation that extract the relative position of instance bounding boxes. Pixels within the human and object bounding boxes take a value of 1 and 0 elsewhere.  A DNN can use this representation to learn 2D filters of human-object spatial configurations. 
Works like those of Liang \et and Gupta \et \cite{liang2020visualsemantic,gupta2018nofrills}, extract relative scale features and relative position features based on the coordinates of the instance bounding boxes. 
As for our work, we extract more nuanced spatial cues from the human pose as illustrated in the left image of Fig. \ref{fig:insight_picture}.

Our relative spatial pose features consist of the coordinate offset between each person's keypoints and the center of (the candidate) object bounding box. We employ He \et pose detector \cite{He_mark_rcnn2017} to estimate 17 keypoints for each person in the 2D image (in COCO \cite{Lin2014MicrosoftContext} format). We define the \textit{i}th human keypoint coordinates as $(x_i, y_i)$ and the relative spatial features $f_{rp}^i$ as:
\begin{equation}
    \label{Eqtn:relative_feature}
    f_{rp}^i:(x_i^{'},y_i^{'}) = (\frac{x_i - x_c^o}{W}, \frac{y_i - y_c^o}{H}).
\end{equation}
where $(x_c^o, y_c^o)$ is the center of object bounding box, and $(W,H)$ is the size of image. We denote the final $17\times2$ relative spatial pose features as $f_{rp} \in \RC^{17 \times 2}$. 

\subsubsection{Absolute Pose Features}\label{subsubsec:abs_pose_feat}
Generally, a person will have different postures when performing different actions. For instance, the human pose when sitting $<$human, sit\_on, chair$>$ or when standing $<$human, stand\_on, chair$>$ are very different. Other times, similar postures may occur when a person acts with different objects (\eg riding a horse or a bicycle). These intuitions indicate that a human's pose intrinsic properties are also useful for HOI detection. 

Similar to \cite{gupta2018nofrills}, we construct absolute keypoint pose features $f_{ap}$ by normalizing with the center of the human bounding box:
\begin{equation}
    \label{absolute_feature}
    f_{ap}^i:(x_i^{''} ,y_i^{''}) = (\frac{x_i}{x_c^h}, \frac{y_i}{y_c^h}).
\end{equation}
where $(x_c^h, y_c^h)$ denotes the center of the human bounding box. We denote the final $17\times2$ dimensional absolute pose features of all keypoints as $f_{ap} \in \RC^{17 \times 2}$.
\subsection{Pose-based Modular Network} \label{subsec:PMN}
An overview of our pose-based module is shown in Fig. \ref{fig:overview}. The module's two streams, project the relative and absolute pose features to higher dimensional features respectively. Then we concatenate and flatten the features before classifying them.

The first stream encodes relative spatial pose features via two fully-connected layers with batch normalization, ReLU activations, and dropout. Eqtn. \ref{eqtn:rel_spatial_pose} defines the operation:
\begin{equation}
    \label{eqtn:rel_spatial_pose}
    h_1 = \mbox{ReLU}(\mbox{ReLU}(f_{sp} \; W_0) \; W_1).
\end{equation}
where $W_0 \in \RC^{2 \times 128}$ and $W_1 \in \RC^{128 \times 64}$ are the learnable weight matrices.

Inspired by \cite{liu2020gastnet, Zhao_2019}, we adopt a GCN \cite{kipf2016semisupervised} layer to process the absolute pose features. We define the human pose as a graph $\mathcal{G = (V,E)}$, where $\mathcal{V}$ is a set of $V$ nodes and $\mathcal{E}$ is a set of $E$ edges. $A \in \RC ^ {V \times V}$ is the adjacent matrix that indicates the connection between joints and $D_{ii} = \sum_{j}A_{ij}$ is a degree matrix. In the second stream, we use a fully-connected layer followed by a GCN layer to process the absolute pose features as indicated in Eqtn. \ref{eqtn:abs_pose_net}:
\begin{equation}
    \label{eqtn:abs_pose_net}
    h_2 = \mbox{ReLU}(\hat{A} \; \mbox{ReLU}(f_{ap} \; W_2) \; W_3).
\end{equation}
where $\hat{A} = \tilde{D}^{-\frac{1}{2}} \tilde{A} \tilde{D}^{-\frac{1}{2}  }$ is the normalized adjacent matrix, with $\tilde{A}= A + I_N$ ($I_N$ is the identity matrix) and $\tilde{D}_{ii} = \sum_{j}\tilde{A}_{ij}$ \cite{kipf2016semisupervised}. Also, $W_2 \in \RC^{2 \times 128}$ and $W_3 \in \RC^{128 \times 64}$ denote the trainable weight matrices.
 
Once the relative and absolute pose features are fed through the network streams, the processed features $h_1$ and $h_2$ are concatenated and fed into a fully-connected layer:
\begin{equation}
    \label{Eqtn:concatenate_layer}
    h  = \mbox{ReLU}((h_1 \otimes h_2) \; W_4).
\end{equation}
where $W_4 \in \RC^{128 \times 64}$.

Later, we reshape the foregoing features: $h \in \RC^{N \times 17 \times 64}$ $\to$ $h^{'} \in \RC^{N \times 1088}$, where $N$ means the number of all box-pairs in each mini-batch. Then we adopt two fully-connected layers to get the final output:
\begin{equation}
    \label{Eqtn:last_layer}
    p_2  = (\mbox{ReLU}(h^{'} \; W_5)) \; W_6.
\end{equation}
where $W_5 \in \RC^{1088 \times 256}$ and $W_6 \in \RC^{256 \times K}$. $K$ denotes the total number of possible actions.

\subsection{Inference and Learning.}\label{subsec:inference}
After passing through the whole framework, we can get the action score factors (the output of last layer) $p_1$ and $p_2$ from the existing HOI detection network branch and our pose-based modular network branch, respectively. In the last inference phase, we directly sum $p_1$ and $p_2$ up, which make our module fully compatible with existing networks. 

As mentioned in Sec. \ref{sec:intro}, HOI detection is a multi-label classification problem, where more than one action/predicate might be assigned to a $<subject, object>$ box-pair. Therefore, We apply a binary sigmoid classifier for each action category:
\begin{equation}
    \label{eqtn:actionScore}
    S^a = sigmoid( p_1 \oplus p_2 ).
\end{equation}
where $S^a \in \RC^{N \times K}$ contains the inferred score for each action categoy for each $<subject, object>$ box-pair.

Our framework is jointly trained end-to-end in a supervised manner by minimizing the multi-label binary cross-entropy loss $BCE(\cdot)$ between inferred the action score $s$ and the ground truth action label $y^{label}$ for each action category:  
\begin{equation}
    \label{Eqtn:training}
    \LC = \frac{1}{N \times K} \sum_{i=1}^N \sum_{j=1}^K  BCE(s_{ij}, y_{ij}^{label})
\end{equation}
See Sec. \ref{subsec:experimental_setup} for more details on training procedures.
\begin{table*}[tb]
\centering
    \begin{tabular}{llccc}
    \toprule
    Method           & Object Detector    & Full(600)$\uparrow$   & Rare(138)$\uparrow$   & Non-Rare(462)$\uparrow$  \\
    \bottomrule
    Shen \textit{et al.} \cite{Shen2018ScalingLearning} & Faster R-CNN with VGG19\cite{Simonyan15_vgg}      & 6.46  & 4.24  & 7.12     \\
    HO-RCNN \cite{chao2018learning}                     & Fast R-CNN\cite{girshick2015fast} & 7.81  & 5.37  & 8.54   \\
    InteractNet \cite{gkioxari2018detecting}            & Faster R-CNN with ResNet-50-FPN   & 9.94  & 7.16  & 10.77  \\
    GPNN \cite{qi2018learning}                          & Deformable ConvNets \cite{Dai2017DeformableNetworks}              & 13.11 & 9.34  & 14.23    \\
    iCAN \cite{gao2018ican}                             & Faster R-CNN with ResNet-50-FPN   & 14.84 & 10.45 & 16.15  \\
    Xu \textit{et al.} \cite{xu2019learning}            & Faster R-CNN with ResNet-50-FPN   & 14.70 & 13.26 & 15.13  \\
    Gupta \textit{et al.} \cite{gupta2018nofrills}      & Faster R-CNN with ResNet-152      & 17.18 & 12.17 & 18.68  \\
    $RP_{T2}C_D$ \cite{Li_2019_CVPR}                    & Faster R-CNN with ResNet-50-FPN   & 17.22 & 13.51 & 18.32  \\
    PMFNet \cite{wan2019pose}                           & Faster R-CNN with ResNet-50-FPN   & 17.46 & 15.65 & 18.00  \\
    Peyre \textit{et al.} \cite{2019PeyreDetecting}     & Faster R-CNN with ResNet-50-FPN   & 19.40 & 14.60 & 20.90  \\
    \toprule
    VS-GATs \cite{liang2020visualsemantic}              & Faster R-CNN with ResNet-50-FPN   & 20.27 &16.03 & 21.54   \\
    VS-GATs + PMN  & Faster R-CNN with ResNet-50-FPN   & \textbf{21.21} & \textbf{17.60} & \textbf{22.29}  \\
    \bottomrule
    \end{tabular}
    \caption{mAP performance comparison with SOTA on the HICO-DET \textit{test} set.}
    \label{tbl:hico_mAP}
\end{table*}

\section{Experiments And Results}\label{sec:exp_results}
In this section, we first describe the experimental datasets and evaluation metrics, followed by more implementation details of our framework (Sec. \ref{subsec:experimental_setup}). Then, we report the quantitative results (Sec. \ref{subsubSec:quantiative_result}) compared with the state-of-the-art methods as well as qualitative detection visualization results (Sec. \ref{subsubSec:qualitative_result}). Finally, we introduce ablation experiments (Sec. \ref{subsec:ablationStudies}) which validate each component of the proposed module.
\subsection{Experimental Setup}\label{subsec:experimental_setup}
\paragraph{\textbf{Datasets.}} 
We adopt two common benchmarks: V-COCO \cite{DBLP:GuptaM15} and HICO-DET \cite{chao2018learning} to evaluate our framework. \textbf{V-COCO} is a subset of the MS-COCO \cite{Lin2014MicrosoftContext} dataset with appended HOI annotations. It contains 10,346 images, where 2,533 form the \textit{train} set, 2,867 form the \textit{val} set, and 4,946 form the \textit{test} set. It contains 16,199 human instances and 29 action annotation categories (five of them have no object interactions (\textit{e.g. smiling}) which we do not consider for HOI detection). \textbf{HICO-DET} is a large-scale dataset which consists of 47,776 images in total (38,118 for training and 9658 for testing). It contains 150K annotated human-object pair instances and 600 HOI categories over 80 object categories (same as \cite{Lin2014MicrosoftContext}) and 117 action categories. The 600 HOI categories are divided into: (i) Full: all 600 categories; (ii) Rare: 138 HOI categories with less than 10 training samples, and (iii) Non-Rare: 462 HOI categories with more than 10 training samples. 
\paragraph{\textbf{Evaluation Metrics.}} 
We adopt the mean average precision (mAP) to measure the detection performance. We consider a detected triplet as true positive when the predicted predicate is correct and both the detected human and object bounding boxes have the intersection-of-union (IoU) $\ge$ 0.5 with respect to the ground truth. 
\paragraph{\textbf{Implementation Details.}}\label{subsec:impldetails}  
We employ Faster R-CNN \cite{renNIPS15fasterrcnn} with a RestNet-50-FPN backbone \cite{he2016deep,lin2017feature} as the object detector. Mask R-CNN \cite{he2017mask} serves as the human pose estimator pre-trained on COCO \cite{Lin2014MicrosoftContext} \footnote{For the object detector and the pose estimator, we directly use Pytorch's  re-implemented API \url{https://pytorch.org/docs/stable/torchvision/models.html}.}. As mentioned in Sec. \ref{subsec:overview}, we choose VS-GATs \cite{liang2020visualsemantic} as the existing HOI detection network in our framework (Fig. \ref{fig:overview}). The architecture of our pose-based modular neural network is illustrated in Fig. \ref{fig:overview}. Note that the object detector, pose estimator and VS-GATs are frozen when training. That's to say, we just train the pose-based module.

We follow the same training scheme from previous works: select the hyperparameters on the \textit{val} set and then retain the model on the \textit{trainval} set (\textit{train} set +\textit{val} set) \footnote{We regard the original training set in HICO-DET as the \textit{trainval} set and follow \cite{liang2020visualsemantic} to split it into the \textit{train} set and the \textit{val} set.}. Following \cite{liang2020visualsemantic}, we set the detection confidence threshold to 0.8 for humans and 0.3 for objects. When training, we use a batch size of 32 and dropout ratio of 0.2. We adopt an Adam optimizer with an initial learning rate of 3e-5. For V-COCO, we reduce the learning rate to 3e-6 at epoch 400 and stop training at epoch 600. For HICO-DET, we reduce the learning rate to 3e-6 at epoch 150 and stop training at epoch 200. We conduct our experiments on a single Quadro P3200 GPU.

\begin{table}[h]
\centering
    \begin{tabular}{lc}
    \toprule
    Method                                                  & $AP_{role}$ (Sce. 1)  \\
    \midrule
    Gupta \textit{et al.} \cite{DBLP:GuptaM15}              & 31.8 \\
    InteractNet \cite{gkioxari2018detecting}                & 40.0 \\
    GPNN \cite{qi2018learning}                              & 44.0 \\
    iCAN \cite{gao2018ican}                                 & 45.3 \\
    Xu \textit{et al.} \cite{xu2019learning}                & 45.9 \\
    Li \textit{et al.} ($RP_DC_D$) \cite{Li_2019_CVPR}      & 47.8 \\
    PMFNet \cite{wan2019pose}                               & 52.0 \\
    \midrule
    VS-GATs  \cite{liang2020visualsemantic}                 & 49.8  \\
    VS-GATs + PMN                                          & \textbf{51.8}  \\ 
    \bottomrule
    \end{tabular}
    \caption{mAP performance comparison with SOTA on the V-COCO \textit{test} set.}
    \label{tbl:vcoco}
\end{table}
\subsection{Results}\label{subsec:result}
\subsubsection{Quantitative Results and Comparisons.} \label{subsubSec:quantiative_result}
Our experiment results (Table \ref{tbl:hico_mAP} and Table \ref{tbl:vcoco}) demonstrate that the proposed Pose-based Modular Network (PMN) combined with VS-GATs \cite{liang2020visualsemantic} beats all SOTA metrics on HICO-DET and achieves comparable result on V-COCO; thus showing the significance of pose cues showing its efficacy and flexibility.

On V-COCO, we achieve an \textbf{51.8 mAP}. Our method improves VS-GATs by \textbf{2 mAP} ($\sim$4.0\%) and also further surpasses most of SOTAs including \cite{Li_2019_CVPR} which also leverage human pose in their network. Note that PMFNet \cite{wan2019pose} considers not only human pose but also human body part features, which make it outperform previous works by a considerable margin. However, our framework still have a comparable performance without the complicated human body part features.
On HICO-DET, our method improves VS-GATs by \textbf{0.98 mAP} ($\sim$4.6\%), \textbf{1.57 mAP} ($\sim$9.8\%), \textbf{0.75 mAP} ($\sim$3.5\%) for the Full, Rare and Non-Rare categories respectively, which makes VS-GATs \cite{liang2020visualsemantic} further outperform existing methods \cite{chao2018learning,gkioxari2018detecting,qi2018learning,gao2018ican,xu2019learning,gupta2018nofrills,Li_2019_CVPR,wan2019pose,2019PeyreDetecting}. 

\subsubsection{Qualitative Results.} \label{subsubSec:qualitative_result}
\begin{figure*}[t]
    \centering
    \includegraphics[width=1\linewidth]{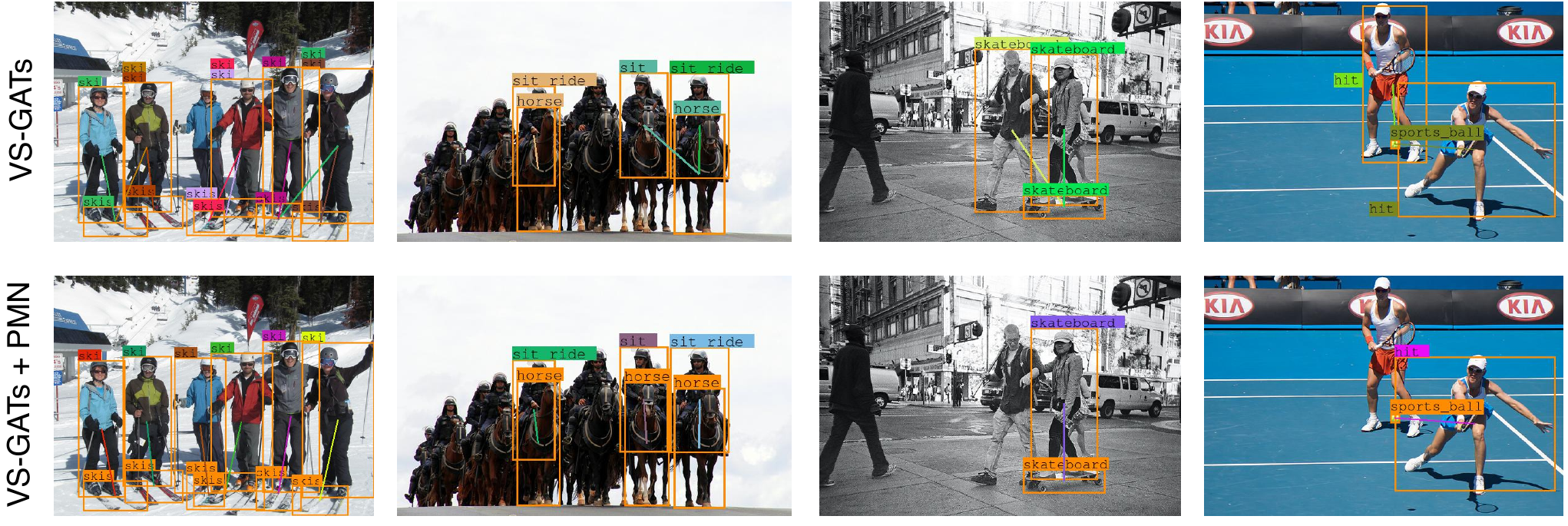}
    \caption{HOI detection results compared with VS-GATs on V-COCO \textit{test} set. The first row is the detection results of original VS-GATs. The second row is the detection results of our framework (VS-GATs + PMN). Subjects and objects are shown in orange bounding boxes.The interaction classes are shown on the subject bounding box and the interactive objects are linked with the line in the same color. We show all triplets whose inferred \textit{action score} exceeds 0.5.}
    \label{fig:qualitative}
\end{figure*}

Fig. \ref{fig:qualitative} shows some Visualization results compared with VS-GATs on V-COCO \textit{test} set. We find that VS-GATs tend to output the false positive detection when multiply persons and objects are close to each other. For example, in the first image, VS-GATs infers the wrong detection that the 2th, 4th, 6th person (from left to right) also ski their neighbors' skis. However, with the proposed pose-based module which explores the detailed spatial cues and intrinsic properties based on human pose, our framework (VS-GATs + PMN) performs better in the crowded scenes as shown in the second row.
\subsection{Ablation Studies} \label{subsec:ablationStudies}
In this section, we perform several ablation studies on HICO-DET. To simplify the training steps, as in \cite{liang2020visualsemantic}, we train the model on the \textit{train} set without further retraining on the \textit{trainval} set.
\paragraph{\textbf{PMN vs. NFPN.}} 
In \cite{gupta2018nofrills}, Gupta \et design their fine-grained layout factor network as a simple three layers MLP to encode the pose features. Similarly, we also construct a No-Frills Pose Network (NFPN) implemented by a a 3-layer MLP with (128,128,117) neurons respectively. The first two layers use batch normalization, ReLU activation, and dropout. We flatten and concatenate our relative spatial and absolute pose features as the 68 (= 17 x 2 + 17 x 2) dimensional input features. From Table \ref{tbl:ablation1}, NFPN also improves VS-GATs but our PMN performs better. 
\begin{table}[h]
\centering
    \begin{tabular}{l|ccc}
    Method              & Full$\uparrow$  & Rare$\uparrow$  & Non-Rare$\uparrow$ \\
    \midrule
    VS-GATs             & 20.27 & 16.03 & 21.54 \\
    VS-GATs + NFPN      & 20.88 & 17.12 & 22.01 \\
    VS-GATs + PMN      & \textbf{21.12} & \textbf{17.59} & \textbf{22.18 } \\
    \end{tabular}
    \caption{\textit{PMN vs. NFPN.} Ablation studies results on HICO-DET \textit{test} set.}
    \label{tbl:ablation1}
\end{table}
\paragraph{\textbf{Relative spatial pose features \vs Absolute pose features.}} 
Table \ref{tbl:ablation2} validates the importance of the pose features in our method. Both set of features facilitate HOI detection and the relative spatial pose features played a more dominant role in this task. 
\begin{table}[h]
\centering
    \begin{tabular}{cc|ccc}
    Relative    &   Absolute    & Full$\uparrow$  & Rare$\uparrow$  & Non-Rare$\uparrow$ \\
    \midrule
    $-$         &   $-$           & 20.27 & 16.03 & 21.54 \\
    $-$         &   $\surd$       & 20.55 & 16.65 & 21.66 \\
    $\surd$     &   $-$           & 20.94 & 16.91 & 21.15 \\
    $\surd$     &   $\surd$       & \textbf{21.12} & \textbf{17.59} & \textbf{22.18} \\
    \end{tabular}
    \caption{\textit{Relative vs. Absolute pose features.} Ablation studies results on HICO-DET \textit{test} set.}
    \label{tbl:ablation2}
\end{table}
\section{Conclusion}\label{sec:conclusion}
In this paper, we propose a pose-based modular network which studies the relative spatial pose feature as well as the absolute pose features to improve HOI detection. The module is easy to combine with existing networks. The experiment results show that our method facilitates the HOI detection system to perform better in the crowded scenes and consistently improves the state-off-the-art method VS-GATs on both V-COCO and HICO-DET benchmarks. 


{
    \small
    \bibliographystyle{ieee_fullname}
    \bibliography{references}
}
\end{document}


\title{Visual-Semantic Graph Attention Network for \\
       Human-Object Interaction Detection \\
       Supplementary Material}
\author{First Author\\
Institution1\\
Institution1 address\\
{\tt\small firstauthor@i1.org}
\and
Second Author\\
Institution2\\
First line of institution2 address\\
{\tt\small secondauthor@i2.org}
}
\maketitle
\begin{abstract}

In this supplementary document, we first provide some qualitative visualization results generated by our proposed method. Then we analyze several false positive detection results and outline our future works. As for the illustration of our code, please refer to the \textbf{readme.md} in code zip file for more details. 

\end{abstract}
\section{Qualitative Results}\label{sec:qualitative_result}
Fig. \ref{fig:qualitative_results} shows some $<subject, predicate,object>$ triplets' detection results on HICO-DET test dataset. From the results, our proposed model is able to detect various kinds of HOIs such as single person interactive with single object, multiple persons interactive with the same object, multiple persons interactive with multiple objects and so on. In particular, consider the first image in second row as an example, our model generates reasonable interaction scores when multiple persons interactive with single object. This, to some extent, suggests that contextual information make the model better to disambiguate the HOIs.      

\section{False Positive Results}\label{sec:false_postive_result}
\begin{figure}[h]
    \centering
    \includegraphics[width=1\linewidth]{fp_img.pdf}
    \caption{Visualization of some false positive HOI detections on HICO-DET testing dataset.Subjects and objects are shown in red bounding boxes.The interaction classes and corresponding triplet's score($S_{R}$) are shown on the subject bounding box and the interactive objects are linked with the same color line. We show all triplets whose inferred action score($S_a$) are greater than 0.3.}
    \label{fig:fp_img}
\end{figure}

In our method, the features we leverage are visual appearance features, spatial features and the word embedding features. We think that the visual appearance features have contained the information of human pose, clothing, human gaze and so on \cite{gkioxari2018detecting}. While in some cases which need fine-grained information such as gaze detection, image depth map, human pose, temporal information, it is still a little bit hard for our model to disambiguate the interaction. The images shown in Fig. \ref{fig:fp_img} are some cases which our model output some wrong detections. 

In image (a), it is hard to infer the right interaction between the $human$ and the $tv$ because we do not provide the model with the fine-grained gaze detection information which is the important cue to handle this situation and there seems no subsidiary relation can provides the model with significant cues. In \cite{gupta2018nofrills}, the author also came across this problem.

In image (b), our model also give a low interaction score between the far away $human$ and the $backpack$ because these two human are overlap and it seems the model need the image depth map information so that it can deal with this case better.

In image (c), the object detector is not able to detect the $pen$ which is not included in dataset and the triple $<human, sign, baseball\_bat>$ is a rare HOI class. Without the significant cues from subsidiary relation in $<human,pen>$, it is a little bit hard to infer the $sign$ action just based on the provided features. In this case, the fine-grained human pose as well as gaze detection information are helpful.

In image (d), this is an challenging case for not only the model but also human ourselves. Our model gave close scores for triplet $<human, catch, frisbee>$ and triplet $<human, throw, frisbee>$ because it is hard to distinguish them without the temporal information.   

\section{Future Works}
Therefore, based on these observations aforementioned, the additional fine-grained human pose, human gaze detection, image depth map information and temporal information will help our model to perform better in some cases. We plan to extend our model with these features in the future.

\begin{figure*}[t]
    \centering
    \includegraphics[width=1\linewidth]{result_img.pdf}
    \caption{Visualization of sample HOI detections on HICO-DET testing dataset. All the triplets are shown with the same conditions described in the caption of Fig. \ref{fig:fp_img}.} 
    \label{fig:qualitative_results}
\end{figure*}

{
    \small
    \bibliographystyle{ieee_fullname}
    \bibliography{references}
}